\documentclass[review]{elsarticle}

\usepackage{amsmath}
\usepackage{lineno,hyperref}
\modulolinenumbers[5]
\usepackage{caption}
\usepackage{subcaption}
\usepackage{nicefrac}
\usepackage{booktabs}
\usepackage[usenames]{xcolor}
\usepackage{units}
\usepackage[acronyms]{glossaries}
\usepackage[ruled,vlined]{algorithm2e}

\journal{Elsevier}

\makeatletter
\providecommand{\doi}[1]{%
  \begingroup
  \let\bibinfo\@secondoftwo
  \urlstyle{rm}%
  \href{http://dx.doi.org/#1}{%
    doi:\discretionary{}{}{}%
    \nolinkurl{#1}%
  }%
  \endgroup
}
\makeatother

\newcommand{\ud}{\,\mathrm{d}}
\newcommand{\pfrac}[2]{\frac{\partial#1}{\partial#2}}

\newcommand{\ol}[1]{\overline{#1}}

\newacronym{dns}{DNS}{direct numerical simulations}
\newacronym{les}{LES}{large eddy simulations}
\newacronym{pdf}{PDF}{probability density function}
\newacronym{pmf}{PMF}{probability mass function}
\newacronym{ml}{ML}{machine learning}
\newacronym{dnn}{DNN}{deep neural network}
\newacronym{dof}{DoFs}{degrees of freedom}
\newacronym{rmse}{RMSE}{root mean square error}
\newacronym{sgs}{SGS}{subgrid-scale}
\newacronym{wale}{WALE}{Wall Adapting Local Eddy-Viscosity}
\newacronym{sgd}{SGD}{stochastic gradient descent}
\makeglossaries
\glsdisablehyper

\begin{document}

\begin{frontmatter}
\title{A block-random algorithm for learning on distributed, heterogeneous data}
\author[ut]{Prakash Mohan}
\ead{prak@utexas.edu}
\author[nrel_hpacf]{Marc T. Henry de Frahan}
\ead{Marc.HenrydeFrahan@nrel.gov}
\author[nrel_cssog]{Ryan King}
\ead{Ryan.King@nrel.gov}
\author[nrel_hpacf]{Ray W. Grout}
\ead{Ray.Grout@nrel.gov}
\address[ut]{Institute for Computational Engineering and Sciences, The University of Texas at Austin, 201 E. 24th Street, POB 4.102, Austin, Texas 78712, USA}
\address[nrel_hpacf]{High Performance Algorithms and Complex Fluids, Computational Science Center, National Renewable Energy Laboratory, 15013 Denver W Pkwy, ESIF301, Golden, CO 80401, USA}
\address[nrel_cssog]{Complex Systems Simulation and Optimization Group, Computational Science Center, National Renewable Energy Laboratory, 15013 Denver W Pkwy, ESIF301, Golden, CO 80401, USA}

\begin{abstract}

  Most deep learning models are based on deep neural networks with
  multiple layers between input and output. The parameters defining
  these layers are initialized using random values and are ``learned''
  from data, typically using stochastic gradient descent based
  algorithms. These algorithms rely on data being randomly shuffled
  before optimization. The randomization of the data prior to
  processing in batches that is formally required for stochastic
  gradient descent algorithm to effectively derive a useful deep
  learning model is expected to be prohibitively expensive for in situ
  model training because of the resulting data communications across
  the processor nodes. We show that the \gls{sgd} algorithm can still
  make useful progress if the batches are defined on a per-processor
  basis and processed in random order even though (i) the batches are
  constructed from data samples from a single class or specific flow
  region, and (ii) the overall data samples are heterogeneous. We
  present block-random gradient descent, a new algorithm that works on
  distributed, heterogeneous data without having to pre-shuffle. This
  algorithm enables in situ learning for exascale simulations. The
  performance of this algorithm is demonstrated on a set of benchmark
  classification models and the construction of a subgrid scale
  \gls{les} model for turbulent channel flow using a data model
  similar to that which will be encountered in exascale simulation.
  
\end{abstract}


\end{frontmatter}

\glsresetall

\section{Introduction}\label{sec:intro}

Simulating complex physics problems while resolving all the relevant
length scales is computationally expensive, requiring millions of core
hours to compute a single realization. Combining \gls{dns} with an
optimization or design cycle is infeasible, creating a need for
reduced-order models. Deep learning is an increasingly popular and
effective modeling technique that use many data to train a neural
network for a variety of tasks~\cite{Lecun2015, Schmidhuber2015,
  Prieto2016, Goodfellow2016, Liu2017}. These tasks range from visual
object recognition and speech recognition to analyzing particle
accelerator data and drug design. Recently, deep learning has been
explored as a tool for creating reduced-order closure models in
turbulent fluid flows~\cite{ling2016reynolds, duraisamy2015new,
  duraisamy2019turbulence}.

For physics simulation, the advent of exascale computing will enable
unprecedentedly high-fidelity simulations. The expectation is to
derive reduced-order models for engineering and design applications
from the many data generated by these simulations. Because it will be
increasingly difficult to save the large amounts of data generated
during the simulations for offline training, this will drive the need
to change existing approaches for training deep learning
models. Online or in situ training, where the model is trained during
the simulation to avoid data storage, has the potential to alleviate
this problem. A data parallel~\cite{zinkevich2010parallelized}
paradigm for deep learning is a practical approach for online
training. In this setting, there are two distinct computational
clusters: one for the physics computations and the other for deep
learning. Data will be transferred from the physics cluster to the
deep learning cluster as needed by the learning algorithms.

\begin{figure}[!tbp]%
  \centering%
  \begin{subfigure}[t]{0.32\textwidth}%
    \includegraphics[width=\textwidth]{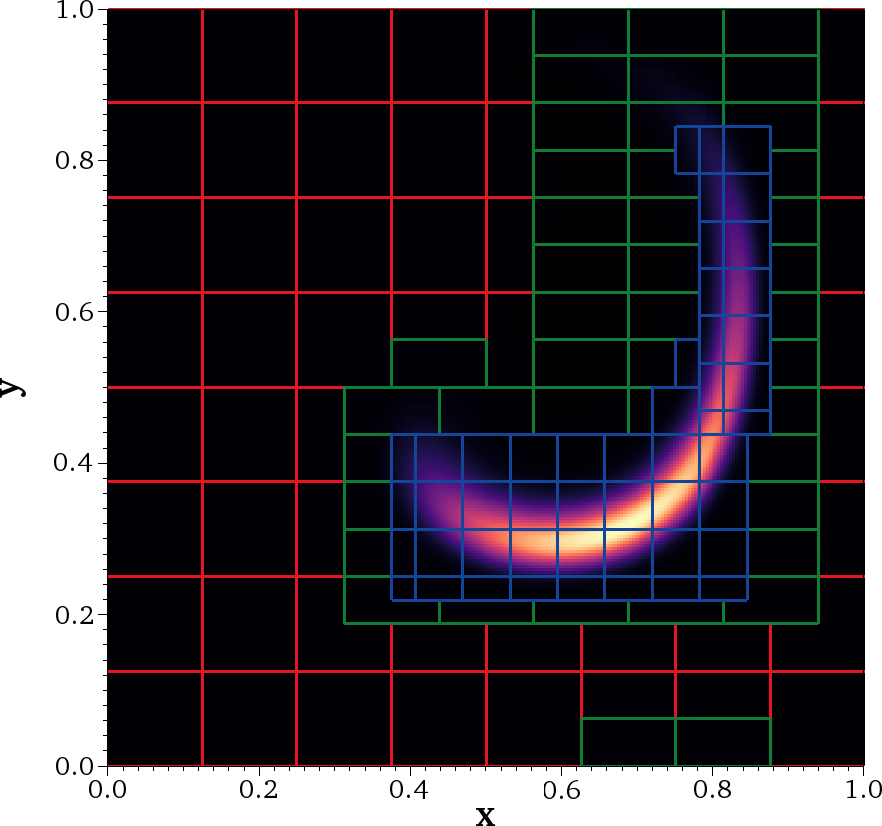}%
    \caption{Advection of a passive scalar. Red: level 0; green: level 1; blue: level 2.}\label{fig:singlevortex}%
  \end{subfigure}\hfill%
  \begin{subfigure}[t]{0.32\textwidth}%
    \includegraphics[width=\textwidth]{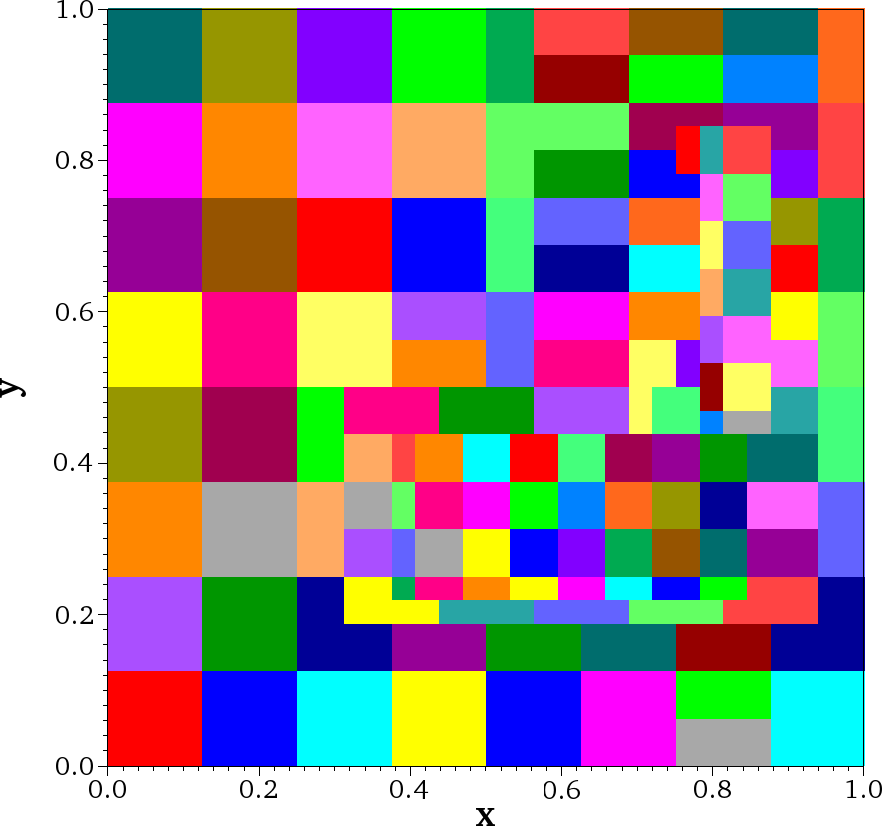}%
    \caption{Memory layout of the simulation data on the different processor ranks. Colors denote the processor ranks.}\label{fig:proc_access}%
  \end{subfigure}\hfill%
  \begin{subfigure}[t]{0.32\textwidth}%
    \includegraphics[width=\textwidth]{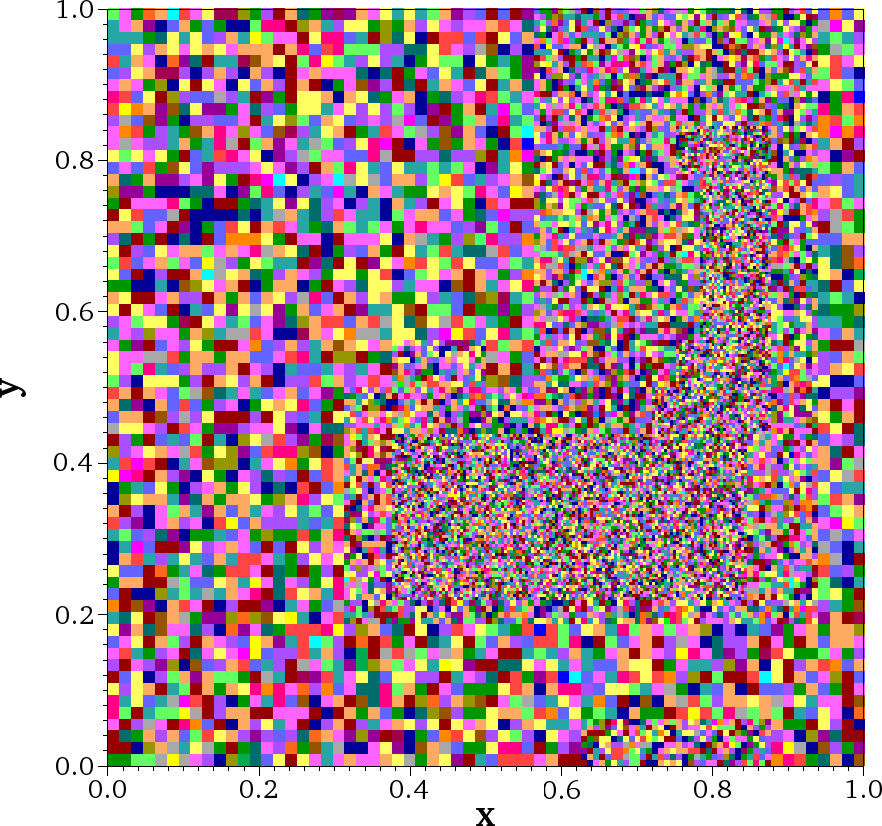}%
    \caption{Fully shuffled memory access for training with \gls{sgd}. Colors denote the different batches.}\label{fig:shuffled_access}%
  \end{subfigure}%
  \caption{Illustration of memory access patterns for the simulation of a vortex using adaptive mesh refinement from the AMReX tutorial.}\label{fig:access_patterns}%
\end{figure}

Most deep learning models use artificial neural networks with multiple
layers to capture nonlinearities. The parameters defining these layers
are ``learned'' from data, typically using algorithms that approximate
gradient descent. With the increase in the amount of available data,
deterministic learning algorithms are often expensive and rarely used
in practice.  \Gls{sgd}~\cite{bottou2010large, Goodfellow2016} and
variants using ``mini-batches'' are commonly used algorithms for practical
learning problems. The stochastic algorithms require the data to be
randomly shuffled~\cite{bottou2010large, Goodfellow2016} for
optimization; however, because fully shuffling the data will be
infeasible for exascale simulations because of the communication costs
of moving the data between processor nodes, the data shuffling
strategy necessary for \gls{sgd} will need to adapt to ensure the
adequate representation of the vastly differing physical processes
occurring in the simulation domain. Shuffling data extracted from a
single computational node will not provide sufficient randomness
because correlations tend to be spatially localized. The randomization
of the data prior to processing in batches that is formally required
for \gls{sgd} to make progress is expected to be prohibitively
expensive for in situ model training. We illustrate the memory
patterns in Figure\,\ref{fig:access_patterns} for the simulation of a
passively advected scalar using adaptive mesh refinement (setup
defined in the AMReX
tutorial\footnote{\url{https://github.com/AMReX-Codes/amrex}}). The
blocks of data are distributed among the different processors and are
heterogeneous, with the mesh adaptivity refining areas of
interest. The memory access pattern for fully shuffling the data for
\gls{sgd} is shown in Figure\,\ref{fig:shuffled_access}. This type of
memory access is detrimental to the simulation performance because the
memory access is uncoalesced, disregards data locality, and requires
many global communications. A single global communication to transfer a large
contiguous chunk of data from one processor is more efficient than many
communications transferring smaller chunks of data from multiple processors. We
show that the \gls{sgd} algorithm can still make useful progress if the batches
are defined on a per-processor basis and processed in random order even though
(i) the batches are constructed from data samples from a single class or
specific flow region, and (ii) the overall data samples are heterogeneous. In
this work, we present a new block-random algorithm that works on distributed,
heterogeneous data without having to pre-shuffle.

This paper is organized as follows. In Section\,\ref{sec:methods}, we
present the problem formulation and describe the proposed methodology
for deep learning of distributed, heterogeneous data. In
Section\,\ref{sec:results}, we detail the architecture of the deep
convolutional neural network used to perform the data recovery process
for fluid flows. In Section\,\ref{sec:benchmarks}, we evaluate our
proposed method on the EMNIST data sets, a standard set of benchmark
problems commonly used in deep learning. In
Section\,\ref{sec:channel}, we apply our methodology to a challenge
problem representative of those encountered for in situ deep learning
in large scale simulations. We construct \gls{sgs} stress models for
large eddy simulations using our proposed methodology and compare it
to standard approaches. Finally, we present conclusions and future work
in Section\,\ref{sec:ccl}.

\section{Methods}\label{sec:methods}
Gradient descent-based algorithms are typical optimization methods used for
training \glspl{dnn}.  Gradient descent is an algorithm to find the set of parameters
$\theta$ that minimize a cost function $J(\theta)$. In the case of \glspl{dnn}, the cost
function is usually a normed distance between the predictions and data in a
training set. The simplest form of gradient descent is:
\begin{equation}
\theta_{k+1} = \theta_k - \eta \nabla_{\theta_k} J(\theta_k)
\label{eq:GD}
\end{equation}
where $\eta$ is the learning rate, $k$ denotes subsequent iterations
of the gradient descent algorithm, and $J$ is computed on the entire
training data set. For very large data sets, this algorithm is very
slow because it computes the cost function on the entire data set for
a single update of the parameters. \Gls{sgd}, in contrast, performs a
parameter update for each sample in the training data set. \gls{sgd}
and \gls{sgd} variants --- such as Adam~\cite{kingma2014adam},
RMSprop~\cite{tieleman2012lecture} and
Adagrad~\cite{duchi2011adaptive} --- have proved to be effective ways
of training \glspl{dnn}. A common addition to \gls{sgd} is to add
``mini-batching'', where the training data are partitioned into
batches of size $n_b$. The batching procedure is used to provide
sequences of approximations of the gradient of the cost function with
respect to the parameters by computing:
\begin{equation}
  \nabla_{\theta_k} J(\theta_k|x \in b, y \in b)
\end{equation}
where $b$ is a batch of training data. Mini-batching provides a better
estimate of $\nabla_\theta J(\theta)$ by using several samples from
the training set instead of only one sample. As a result, the
parameter updates tend to be less noisy. It is also computationally
more efficient by using vectorized computations and parallelism
provided by modern architectures. In practice, the shuffled training
data is divided into batches, the batches are then randomly shuffled
before each pass through the training data, and each batch is used to
provide gradient approximations to update the neural network model
parameters. It has been shown that the \gls{sgd} gradient
approximations converge to the true gradient in
expectation~\cite{bottou2010large}.

Being stochastic, however, these algorithms require the data to be randomly
shuffled to converge to a minimum of the cost function. Results from
Section \ref{sec:benchmarks} show how these algorithms fail to converge without
shuffling when the batches have inherent bias.  As described in Section
\ref{sec:intro}, this shuffling operation is infeasible for online learning on
exascale simulations. We propose a block-random algorithm for use in these
cases where the data ordering needs to remain unchanged. The algorithm operates
by swapping the order of shuffling and batching operations. This shuffling of
batches appears to be sufficient for learning the parameters even when the data
are highly ordered, resulting in batches with high bias. Although individual
batches have high bias, this shuffling operation ensures that the same bias is
not seen by the optimizer consecutively, enabling it to still get to a local
minima of the cost function. This behavior will be shown over a variety of
benchmark problems in Section \ref{sec:results}. In a distributed data setting,
the shuffling of batches will be achieved by picking a random block of data and
getting a batch of size $n_b$ from it.

\begin{algorithm}[h]
\DontPrintSemicolon
    \textbf{Parameters:} learning rate $\eta$, batch size $n_b$, number of epochs $n_e$\;
    \KwIn{training data with N samples}
    \SetAlgoLined
    \While{$i \leq n_e$}{
        randomly shuffle data\;
        partition data into mini batches $B_k$ of size $n_b$\;
        \While{$k \leq \nicefrac{N}{n_b}$}{
          $\theta_{k+1} = \theta_k - \eta \nabla_{\theta_k} J(\theta_k|B_k)$\;
          $k = k +1$\;
        }
        $i = i +1$\;
    }
    \caption{Stochastic gradient descent with mini-batching}\label{algo:mini-batch}
\end{algorithm}

\begin{algorithm}[H]
    \DontPrintSemicolon
    \textbf{Parameters:} learning rate $\eta$, batch size $n_b$, number of epochs $n_e$\;
    \KwIn{training data with N samples}
    \SetAlgoLined
    partition data into batches $B_k$ of size $n_b$\;
    \While{$i \leq n_e$}{
        \While{$k \leq \nicefrac{N}{n_b}$}{
          $\theta_{k+1} = \theta_k - \eta \nabla_{\theta_k} J(\theta_k|B_k)$\;
          $k = k +1$\;
        }
        $i = i +1$\;
    }
    \caption{Block-unshuffled gradient descent}\label{algo:block-unshuffled}
\end{algorithm}

\begin{algorithm}[H]
    \DontPrintSemicolon
    \textbf{Parameters:} learning rate $\eta$, batch size $n_b$, number of epochs $n_e$\;
    \KwIn{training data with N samples}
    \SetAlgoLined
    partition data into mini batches $B_k$ of size $n_b$\;
    \While{$i \leq n_e$}{
        randomly shuffle ordering of batches $B_k$\;
        \While{$k \leq \nicefrac{N}{n_b}$}{
            $\theta_{k+1} = \theta_k - \eta \nabla_{\theta_k} J(\theta_k|B_k)$\;
            $k = k +1$\;
        }
        $i = i +1$\;
    }
    \caption{Block-random gradient descent}\label{algo:block-random}
\end{algorithm}

Throughout this work, a batch denotes the data set that is used by the
\gls{sgd} algorithm to evaluate the model and perform the
back-propagation of the neural network weights. A block, or a class
for the image classification benchmark problems, denotes a homogeneous
data set that is distributed among the multiple processors. An epoch
consists of training the model on the entire training data set.

The relative performance of the algorithms will be tested on a suite
of data sets with the results shown in Section \ref{sec:results}. Each
data set will be tested in three different scenarios: (i)
\textit{shuffled} --- fully shuffling all the data and then creating
the batches (Algorithm\,\ref{algo:mini-batch}); (ii)
\textit{block-unshuffled} --- arranging the data in blocks to
emphasize bias and running it without shuffling
(Algorithm\,\ref{algo:block-unshuffled}); (iii) \textit{block-random}
--- accessing the arranged blocks in a block-random fashion
(Algorithm\,\ref{algo:block-random}). For the
\textit{block-unshuffled} and \textit{block-random} cases, we pick the
worst-case scenario for the bias. For instance, in the image
classification case, each block (and consequently each batch) will
contain only one class as shown in Table\,\ref{table:illustration}.

\begin{table}[!htb]
    \begin{subtable}{.3\linewidth}
      \centering
        \begin{tabular}{ | c | c | c | c | c | }
        \hline
            4 & 0 & 2 & \ldots & 1 \\
            3 & 2 & 9 & \ldots & 6 \\
            8 & 7 & 1 & \ldots & 8 \\
            9 & 8 & 4 & \ldots & 5 \\
            0 & 5 & 5 & \ldots & 7 \\
            6 & 1 & 3 & \ldots & 2 \\
            5 & 3 & 6 & \ldots & 9 \\
          \vdots & \vdots& \vdots & \ldots & \vdots  \\
            1 & 3 & 5 & \ldots & 7 \\
        \hline
        \end{tabular}
        \caption{\it shuffled}
    \end{subtable}%
    \begin{subtable}{.4\linewidth}
      \centering
        \begin{tabular}{ | c | c | c | c | c | c | }
        \hline
            0 & 0 & \ldots & 1 & \ldots & 9 \\
            0 & 0 & \ldots & 1 & \ldots & 9 \\
            0 & 0 & \ldots & 1 & \ldots & 9 \\
            0 & 0 & \ldots & 1 & \ldots & 9 \\
            0 & 0 & \ldots & 1 & \ldots & 9 \\
            0 & 0 & \ldots & 1 & \ldots & 9 \\
            0 & 0 & \ldots & 1 & \ldots & 9 \\
            \vdots & \vdots& \vdots & \vdots &  \vdots & \vdots  \\
            0 & 0 & \ldots & 1 & \ldots & 9 \\
        \hline
        \end{tabular}
        \caption{\it block-unshuffled}
    \end{subtable}%
    \begin{subtable}{.4\linewidth}
      \centering
        \begin{tabular}{ | c | c | c | c | c | }
        \hline
            4 & 3 & 8 & 0 & \ldots \\
            4 & 3 & 8 & 0 & \ldots \\
            4 & 3 & 8 & 0 & \ldots \\
            4 & 3 & 8 & 0 & \ldots \\
            4 & 3 & 8 & 0 & \ldots \\
            4 & 3 & 8 & 0 & \ldots \\
            4 & 3 & 8 & 0 & \ldots \\
            \vdots & \vdots& \vdots & \vdots &  \vdots \\
            4 & 3 & 8 & 0 & \ldots \\
        \hline
        \end{tabular}
        \caption{\it block-random}
    \end{subtable}
    \caption{
    Illustration of batches from the three scenarios for training on EMNIST digits
    data set. Each column represents a single batch, in the order processed by the
    algorithms. The \textit{shuffled} case has no bias across the batches, but both
    \textit{block-unshuffled} and \textit{block-random} just have one class in each
    batch. The main difference between \textit{block-unshuffled} and
    \textit{block-random} is that the consecutive batches are not from the same
    class in the latter.}
    \label{table:illustration}
\end{table}

\section{Results}\label{sec:results}

\subsection{Benchmark results on EMNIST data sets}\label{sec:benchmarks}

To benchmark the different learning algorithms, we use the EMNIST data
sets~\cite{Cohen2017}. This is a commonly used data set of
$28 \times 28$ pixel images of handwritten character letters and
digits. We trained models using the different training algorithms
presented in Section\,\ref{sec:methods} for the seven different data
sets: ``fashion'', ``digits'', ``letters'', ``byclass'', ``balanced'',
and ``mnist''.

\subsubsection{Neural network architecture}
The neural network architecture is two
fully connected hidden layers each comprising 512 nodes, a rectified
linear unit activation function, and a dropout layer with a dropout
rate of $0.2$. The final layer includes a softmax activation function
for the category probabilities:
\begin{align}
  y = S(x) = \frac{\exp{(x)}}{\sum^n_{i=1} \exp{(x_i)}},
\end{align}
where $x$ is the layer input vector of size $n$, and $y$ is the layer
output vector of size $n$, on the output layer to ensure that
$\sum^n_{i=1} y_i = 1$ and $y_i \in [0,1]~\forall i = 1, \dots,
n$. The loss function is the categorical cross-entropy loss, and the
\gls{sgd} algorithm for this work is the Adam
optimizer~\cite{Kingma2014}. The learning procedure occurred over 50
epochs, where one epoch consists of training the model on the entire
training data set. The deep learning framework was implemented through
Keras~\cite{Chollet2015} with the TensorFlow
backend~\cite{tensorflow2015-whitepaper}.

\subsubsection{Assessments of learning algorithm performance}

\begin{figure}[!tbp]%
  \centering%
  \begin{subfigure}[t]{0.48\textwidth}%
    \includegraphics[page=2, width=\textwidth]{benchmark_summary.pdf}%
    \caption{Model accuracy ($n_b=64$) for the three learning algorithms. Red: shuffled; green: sorted by class; blue: block-random.}\label{fig:sum0}%
  \end{subfigure}\hfill%
  \begin{subfigure}[t]{0.48\textwidth}%
    \includegraphics[page=1, width=\textwidth]{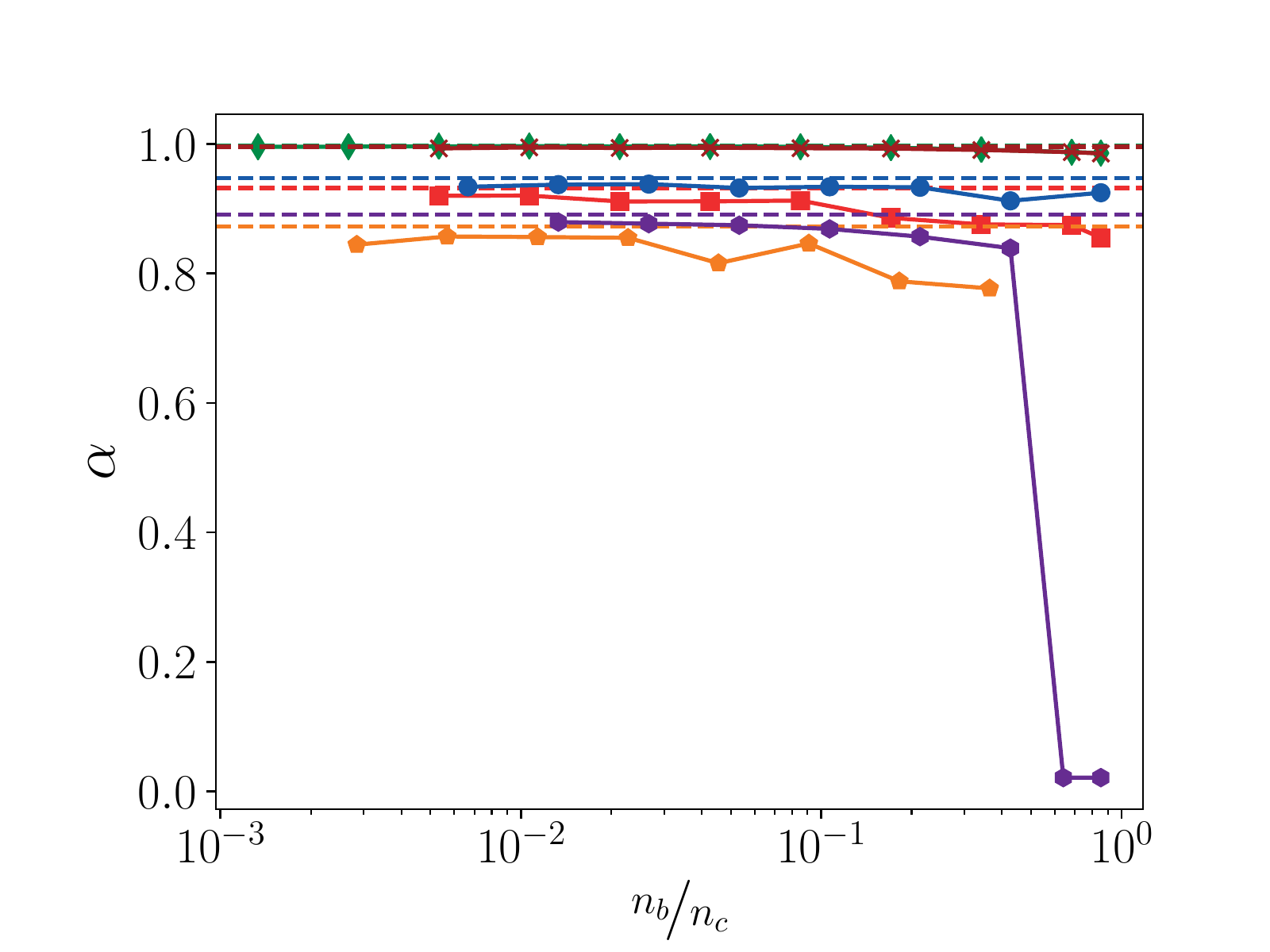}%
    \caption{Model accuracy of block-random algorithm as a function of $\nicefrac{n_b}{n_c}$. Red squares: ``fashion''; green diamonds: ``digits''; blue circles: ``letters''; orange pentagon: ``byclass''; purple hexagon: ``balanced''; burgundy crosses: ``mnist''.}\label{fig:sum1}%
  \end{subfigure}%
  \caption{Model accuracy, $\alpha = \nicefrac{\text{number of correctly classified images}}{\text{total number of images}}$, on validation data set for different benchmark cases.}\label{fig:predictions}%
\end{figure}

Figure\,\ref{fig:sum0} shows the model accuracy using the three
different learning strategies. The results indicate that the block-random
algorithm performs as well as the shuffled algorithm with little
difference in the model accuracy. The block-unshuffled case performs poorly
for all benchmark cases. Additionally, we investigated the effect of
the ratio $\nicefrac{n_b}{n_c}$, where $n_b$ is the batch size, and
$n_c$ is the number of samples in each class, i.e., a block of
homogeneous data, Figure\,\ref{fig:sum1}. The model accuracy for the
block-random algorithm decreases as a function of the ratio
$\nicefrac{n_b}{n_c}$. This is because as this ratio increases, the
\gls{sgd} algorithm operates on batches with little
class variation.

\subsection{Channel flow}\label{sec:channel}

To evaluate the performance of the algorithm in an exascale-like setting, we
developed a \gls{dnn} model for the ``closure problem'' for
\gls{les} in computational fluid dynamics. \Gls{dns} of turbulent
flows, in which all the physical length scales are resolved
explicitly, require large computational resources and are often
unfeasible for engineering and design applications. \Gls{les}
alleviate the computational requirements by resolving the large-scale
motions and modeling the \gls{sgs}, i.e., the length scales that
are not resolved by the discretization grid. In computational fluid
dynamics, \gls{les} solve the filtered Navier-Stokes equations,
presented here in their incompressible form:
\begin{align}
  \label{eq:fns}
  \pfrac{\ol{u}_i}{t} + \pfrac{}{x_j} \left( \ol{u}_i \ol{u}_i \right) &= \pfrac{}{x_j} \left( \nu \pfrac{\ol{u}_i}{x_j} \right) - \frac{1}{\rho} \pfrac{\ol{p}}{x_i} - \pfrac{\tau_{ij}}{x_j}\\
  \pfrac{\ol{u}_i}{x_i} &= 0
\end{align}
where $i$ and $j = 1,2,3$; $x_i$ is the coordinate; $u_i$ is the
velocity in the $x_i$ direction; $p$ is the pressure; $\nu$ is the
kinematic viscosity; $\ol{\cdot}$ is the filtering operation, defined
for an variable $\phi$ as
$\ol{\phi} = \int_{\mathcal{D}} \phi(x) G(x,x') \ud{} x'$, where
$G(x,x')$ is a filter function and $\mathcal{D}$ is the domain; and
$\tau_{ij}$ is the \gls{sgs} stress defined as
$\tau_{ij} = \ol{u_i u_j} - \ol{u}_i\ol{u}_j$. The \gls{les} system of
equations is unclosed because of the \gls{sgs} stress term and
requires a model for the \gls{sgs}. Extensive work has been done to
determine appropriate models for the \gls{sgs}
stress~\cite{Rogallo1984,Lesieur1996,Piomelli1999,Meneveau2000}. For
example, an early approach~\cite{Smagorinsky1963} uses an eddy
viscosity closure that relates resolved velocity gradients to the
\gls{sgs} stress according to:
\begin{align}
  \label{eq:smag}
  \tau_{ij} = -2 (C_s \ol{\Delta})^2 |\ol{S}| \ol{S}_{ij}
\end{align}
where
$\ol{S}_{ij} = \nicefrac{1}{2} \left( \pfrac{\ol{u}_i}{x_j} +
  \pfrac{\ol{u}_j}{x_i} \right)$,
$|\ol{S}| = \sqrt{2 \ol{S}_{ij} \ol{S}_{ij}}$, $\ol{\Delta}$ is the
filter length scale, and $C_s$ is a constant determined through the
\gls{dns} of turbulent flows. Though \gls{sgs} models have received
much attention, because they are often tuned to simple configurations,
the accuracy of these models continues to be problematic in a wide
range of flows. In this section, we will use deep learning to
construct a \gls{sgs} stress model for $\tau_{12}$.

\begin{figure}[!tbp]%
  \centering%
  \reflectbox{\includegraphics[width=0.5\textwidth]{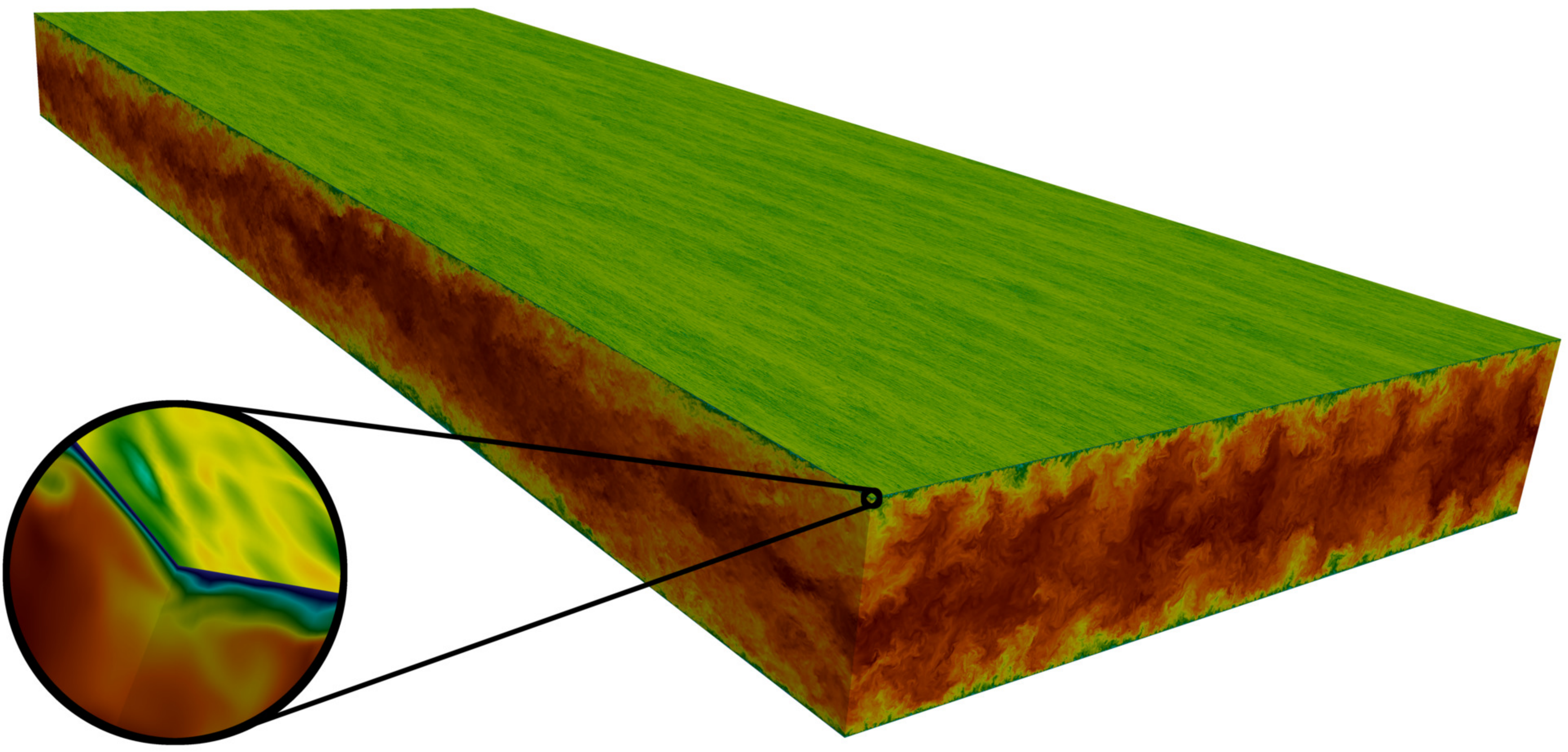}}%
  \caption{Streamwise velocity and wall-shear stress (inset) in the turbulent channel flow \gls{dns}. Figure generated by Dr.~Myoungkyu Lee, using data from~\cite{lee2015direct}}%
  \label{fig:dns_channel}%
\end{figure}

We emphasize that the objective of this work is not to derive the most
accurate \gls{sgs} model for turbulence, which has been a focus of
recent investigations using deep learning~\cite{Ling2016, Maulik2017};
rather, it is to illustrate how to use \gls{dns} data from exascale-like
simulations to develop an accurate model in the context of
distributed, heterogeneous data. As such, we will use the \gls{dns} of
an incompressible channel flow at a friction Reynolds number ($Re_\tau$) of 5186
by \citet{lee2015direct}. The simulations were performed using the code
PoongBack~\cite{lee2013petascale,lee2014experiences}, with 242 billion degrees
of freedom (10240 in $x$, 1536 in $y$, and 7680 in $z$), and was run on 52488 cores,
using approximately 400 million core hours of computation. The incompressible
channel flow exhibits high inhomogeneity and anisotropy, as shown in
Figure\,\ref{fig:dns_channel}, from the presence of the walls.  This makes it a
challenging test case for the block-random algorithm because there will be a high
bias between data from each spatial block.

\subsubsection{Data generation process}

To get data for constructing a \gls{sgs} model, first the nonlinear $u_i u_j$
terms are computed from the \gls{dns} velocity fields. A circular Fourier
cutoff filter is then applied to the resulting fields, and $\tau_{ij}$ is
computed from the filtered fields.  The cutoff wavelength is chosen as
$\lambda^+\approx1500$ in the wall-parallel directions using insights from
\citet{lee_moser_2019}. This results in a grid of dimensions $(200,1536,128)$ in $(x,y,z)$
with more than 39 million data points. The input model variables are the three
filtered velocities and nine filtered velocity gradients. The model is
therefore learning a pointwise functional form:
\begin{align}
  \label{eq:nn_function_model}
  \tau_{ij}|_k = f\left( \ol{u}_i|_k, \left.\pfrac{\ol{u}_i}{x_j}\right|_k \right)
\end{align}
where $k$ denotes a point in the domain. ,The resulting data are then
arranged into spatial blocks of size $(16,16,16)$ in $(x,y,z)$, with
each block representing a computational node. In this setting, there
will be a total of 9600 nodes containing data, which mimics a
distributed large-scale computation.

\subsubsection{Neural network architecture}

The neural network architecture is a feed-forward, fully connected
\gls{dnn}. The hidden layers each comprise fully connected
nodes. The first hidden layer contains a leaky rectified linear unit
activation function:
\begin{align}
  \label{eq:relu}
  y = R(x) =
  \begin{cases}
    x, & \text{ if } x \geq 0, \\
    \gamma x, & \text{ otherwise, }
  \end{cases}
\end{align}
where $x$ is the layer input vector, $y$ is the layer output vector,
and $\gamma=10^{-2}$ is a small slope. The other hidden layers contain
a hyperbolic tangent activation function. The final network layer does
not contain an activation function. The loss used to train the network
is a mean squared error loss function. The specific \gls{sgd} algorithm for this work is the Adam
optimizer~\cite{Kingma2014} because it presents many more advantages
than traditional \gls{sgd} by maintaining a
per-parameter learning rate, which is adapted during training based on
exponential moving averages of the first and second moments of the
gradients. The deep learning framework was implemented through
Keras~\cite{Chollet2015} with the TensorFlow
backend~\cite{tensorflow2015-whitepaper}. To find a reasonably
accurate model for this study, a sweep of the model's hyperparameters
was performed, exploring the following combinations: the initial
learning rate was varied from $10^{-2}$ to $10^{-5}$; the number of
layers, $L$, was varied from 2 to 16; and the number of nodes in
each layer was varied from 8 to 512. The sweeps were performed on the
1 million samples from the fully shuffled training data set. A
neural network comprising an initial learning rate of $10^{-4}$,
four hidden layers, and 128 nodes in each layer (for a total of 68000
trainable parameters) led to a model that was accurate without
necessitating more than eight hours of training on an Intel Skylake
workstation. This set of model hyperparameters is used in all
subsequent results.

\subsubsection{Assessments of learning algorithm performance}

In this section, we present the results of the three different
learning algorithms presented in this work: (i) fully shuffling all
the data; (ii) using the data as they are in the high performance
computing simulation (block-unshuffled), (iii) accessing the data in
block-random fashion, as discussed in Section\,\ref{sec:methods}. Model
quantities are denoted by superscript $\cdot^m$, and quantities
computed with respect to the training and validation data sets are
subscripted with $t$ and $v$, respectively. The mean squared error is defined as
$\epsilon = \frac{1}{|\mathcal{D}|} \sum_{i\in \mathcal{D}} (\tau_{12}
- \tau_{12}^m)^2$, where $\mathcal{D}$ denotes the data set, and
$|\cdot|$ is the cardinality of the set. The physics model used for
comparisons is the \gls{wale} model~\cite{Ducros1998}, a model
specifically designed for \gls{les} of wall-bounded flows.

\begin{figure}[!tbp]%
  \centering%
  \begin{subfigure}[t]{0.48\textwidth}%
    \includegraphics[page=1, width=\textwidth]{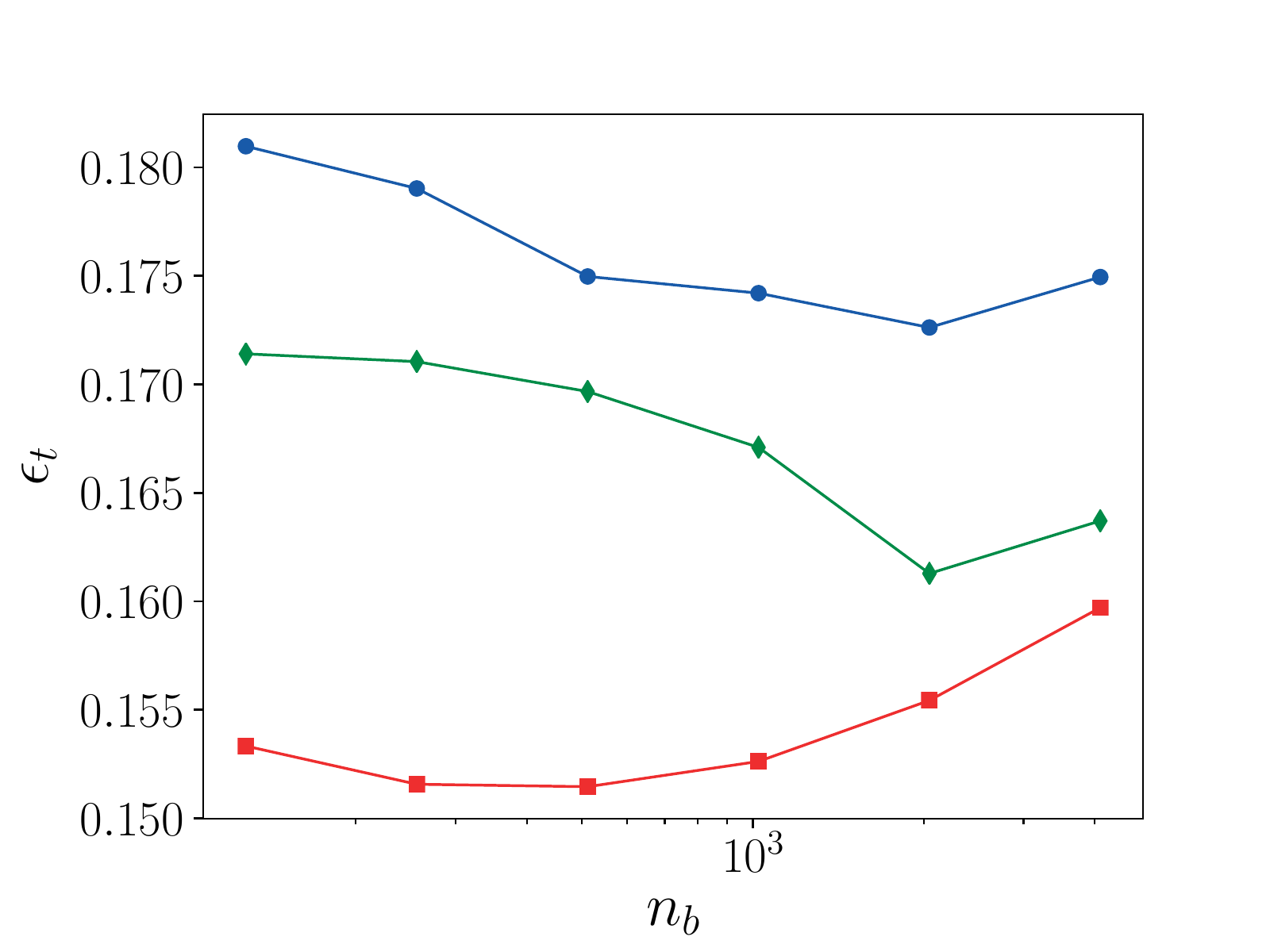}%
    \caption{Training error, $\epsilon_t$.}%
  \end{subfigure}\hfill%
  \begin{subfigure}[t]{0.48\textwidth}%
    \includegraphics[page=2, width=\textwidth]{accuracies_batchsize.pdf}%
    \caption{Validation error, $\epsilon_v$.}%
  \end{subfigure}%
  \caption{Mean squared error for the three different learning algorithms as a function of the batch size. Red squares: shuffled; green diamonds: block-unshuffled; blue circles: block-random.}\label{fig:accuracies_batchsize}%
\end{figure}

The training and validation mean squared error, $\epsilon_t$ and
$\epsilon_v$, respectively, are shown in
Figure\,\ref{fig:accuracies_batchsize} for the three different
learning algorithms as a function of the batch size, $n_b$. The
training error for the shuffled case reaches a minimum at $n_b=256$
and increases for higher $n_b$. The validation error, however, remains
constant and smaller than the other two algorithms. The training error
for the block-unshuffled algorithm is less than that for the block-random
algorithm, though the validation error is three times larger than the
other algorithms. This indicates that the model is capturing the
batches of data toward the end of the training iteration but fails to
adequately represent the full range of data. The training error
decreases as a function of batch size because it is able to get a more
representative data batch. This results, however, in an increasing
validation error because it overfits the data available at the end of the
training iteration. The block-random algorithms exhibits a validation
error that is approximately $10\%$ higher than the shuffled algorithm,
and it remains small as the batch size increases. For the remaining
results presented in this section, the batch size is fixed at 256.

The normalized \gls{pdf} of the error and the conditional means of the
predictions are shown in Figure\,\ref{fig:predictions}. The shuffled
algorithm exhibits a sharp \gls{pdf} of the error and a conditional
means of the predictions identical to the filtered \gls{dns}
throughout the channel domain. The block-random algorithm has a
similar error \gls{pdf} but underpredicts the peak $\tau_{12}$ by
approximately $8\%$. The block-unshuffled algorithm fails to capture the
conditional means throughout the channel. The \gls{wale} physics model
overpredicts the peak $\tau_{12}$ by a factor of two and predicts that
the peak occurs closer to the channel centerline. The \gls{wale} error
\gls{pdf} has a higher variance than the shuffled and block-random
algorithms.

\begin{figure}[!tbp]%
  \centering%
  \begin{subfigure}[t]{0.48\textwidth}%
    \includegraphics[page=4, width=\textwidth]{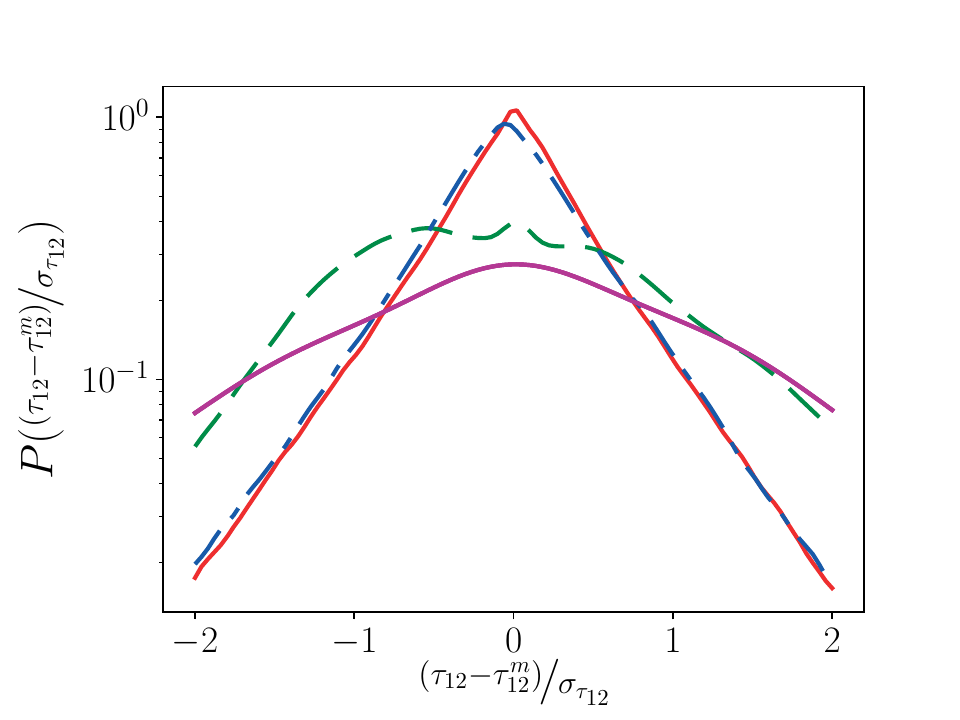}%
    \caption{Probability density function of the error normalized by the standard deviation of $\tau_{12}$.}%
  \end{subfigure}\hfill%
  \begin{subfigure}[t]{0.48\textwidth}%
    \includegraphics[page=17, width=\textwidth]{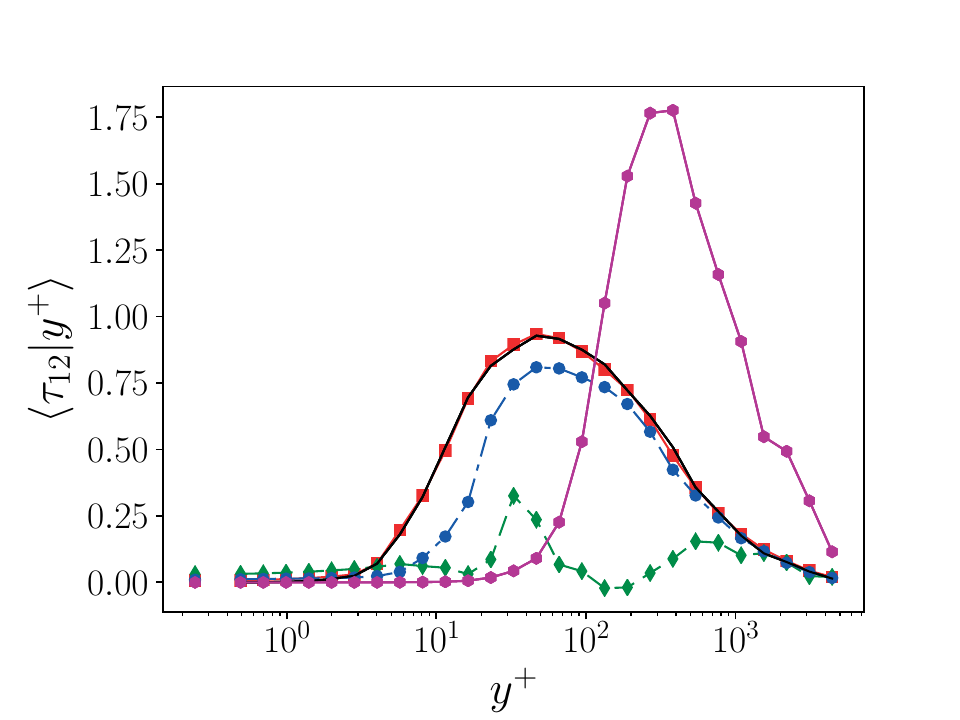}%
    \caption{Conditional means as a function of wall distance, $y^+ = Re_\tau y$.}%
  \end{subfigure}%
  \caption{Model accuracy on validation data set. Red squares and solid: shuffled; green diamonds and dashed: block-unshuffled; blue circles and dash-dotted: block-random; purple hexagons and solid: \gls{wale} physics model; black solid: filtered \gls{dns}}\label{fig:predictions}%
\end{figure}%

As evidenced by these results, the key factor determining the
performance of the model is the order of the blocks used by the
\gls{sgd} algorithm to adjust the model
parameters. To quantify the difference between the learning
algorithms, we use the Jensen-Shannon divergence~\cite{Endres2003,
  Osterreicher2003}, a measure of the similarity between two
\glspl{pdf}. It is a symmetric version of the Kullback-Leibler
divergence~\cite{Kullback1987}, and it is defined as:
\begin{align}
  \label{eq:jsd}
  J(Q,R) = \frac{1}{2} \left( D(Q , M) + D(R , M)\right)
\end{align}
where
$D(Q,R) = \sum_{i=1}^n R(i) \ln{\left( \frac{R(i)}{Q(i)} \right)}$;
$M = \nicefrac{1}{2} \left( Q+R \right)$; $Q$ and $R$ are \glspl{pdf}
of length $n$; and $0\leq J(Q,R) \leq \ln{(2)}$. Low values indicate
more similarity between $Q$ and $R$. The Jensen-Shannon divergence has
several advantages compared to the Kullback-Leibler divergence:
symmetry, i.e., $J(Q,R) = J(R,Q)$, bounded, and variable
support. This metric is used to quantify the difference between the
different \glspl{pdf}: the \gls{pdf} of $\tau_{12}$ in the validation
data set, $Y = P(\tau_{12} \in \mathcal{D}_v)$, and the \gls{pdf} of a
given batch, $Y_b = P(\tau_{12} \in \mathcal{D}_b)$, where
$\mathcal{D}_b$ is the set of data in batch $b$. We compare (i) how
the data in each batch are representative of the data in the entire
domain by computing $J(Y, Y_b)$ for all the batches in an epoch and
(ii) how the data in each batch vary compared to the previous batch
by computing $J(Y_{b-1}, Y_b)$.

\begin{figure}[!tbp]%
  \centering%
  \begin{subfigure}[t]{0.48\textwidth}%
    \includegraphics[page=2, width=\textwidth]{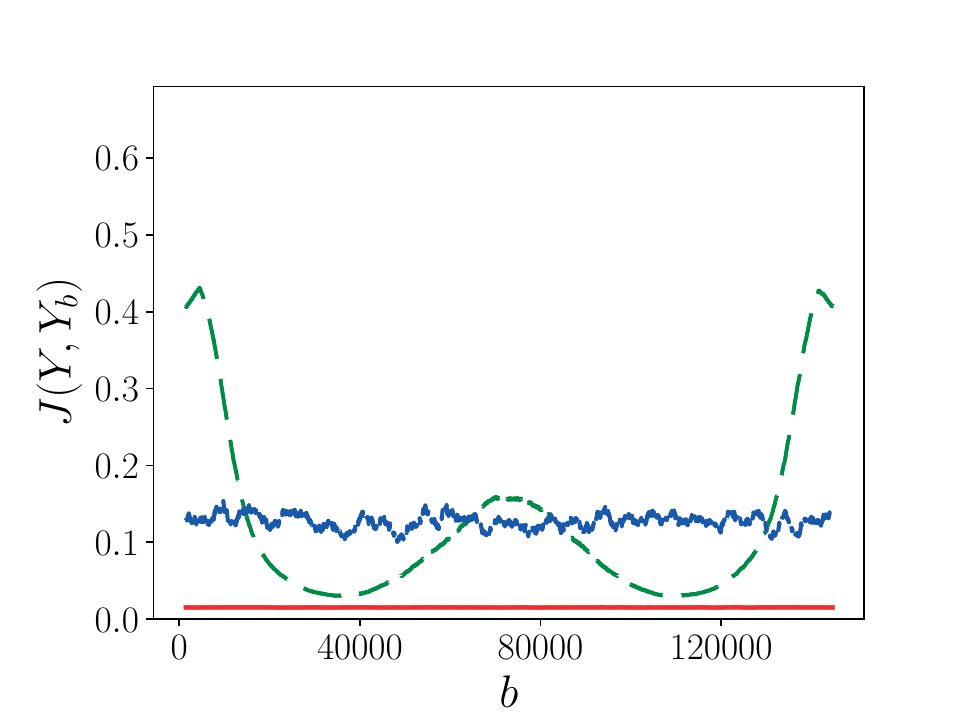}%
    \caption{Comparison between each batch and the entire data set.}%
  \end{subfigure}\hfill%
  \begin{subfigure}[t]{0.48\textwidth}%
    \includegraphics[page=5, width=\textwidth]{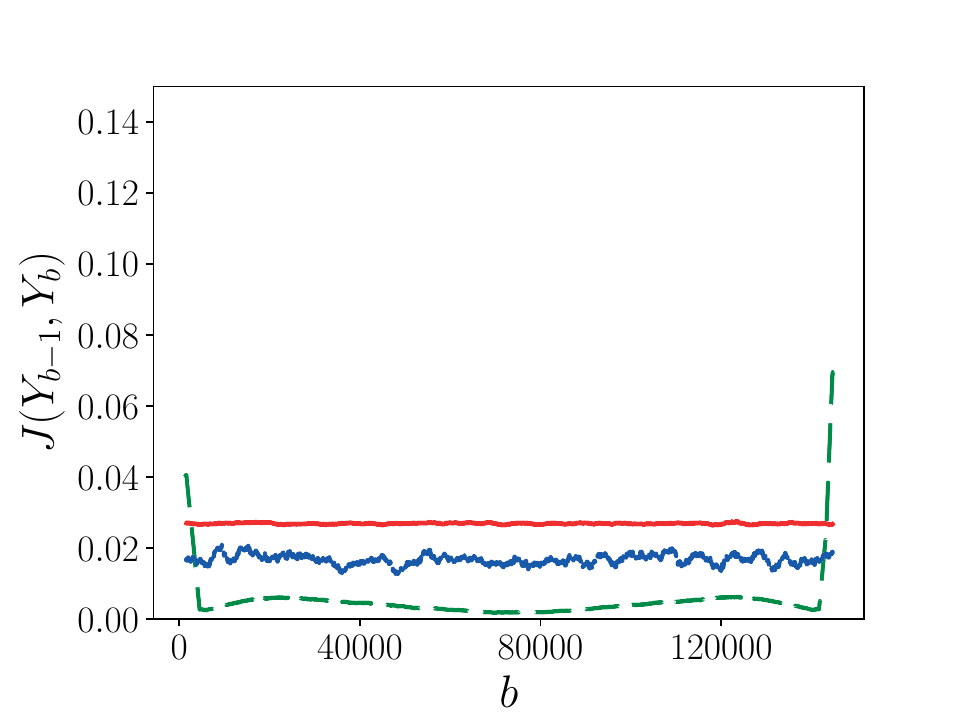}%
    \caption{Comparison between subsequent batches.}%
  \end{subfigure}%
  \caption{Characterizing the differences between batches. Solid red: shuffled; dashed green: block-unshuffled; dash-dotted blue: block-random.}\label{fig:jsds}%
\end{figure}

Figure\,\ref{fig:jsds} illustrates the different metrics for the three
different learning algorithms. For the fully shuffled case, each batch
exhibits a \gls{pdf} similar to that of the entire data set. The
difference between $Y$ and $Y_b$ for the block-random case is higher
but remains constant during the training epoch. For the block-unshuffled
case, there is a clear structure to $J(Y,Y_b)$ because of the
heterogeneity of the data near the channel walls (beginning and ending
of the training epoch). For all three algorithms, the difference
between each subsequent batch is negligible. These results indicate
that ensuring that $J(Y,Y_b)$ remains less than 0.2 and constant
throughout the training epoch is a criteria for achieving high model
accuracy.

\section{Conclusions}\label{sec:ccl}

Effectively training \gls{dnn} models often assumes that the \gls{sgd}
algorithm processes data in batches of data that have been randomized
prior to model training. This is expected to be prohibitively
expensive for in situ model training from the perspective of
communication between computing nodes, as illustrated in
Figure\,\ref{fig:access_patterns}. We showed that the \gls{sgd}
algorithm can still train an effective model if the batches are
processed in random order even if the batches are each comprised of
similar data. In this work, we demonstrated a block-random learning
algorithm for training \glspl{dnn} in the context of distributed
heterogeneous data for data parallelism learning, a situation that
will be increasingly common as the exascale era approaches and models
are trained in situ. The block-random learning algorithm was tested on
several different cases. For the benchmark EMNIST data sets, the
block-random algorithm achieves accuracy similar to the traditional
fully shuffled learning algorithm. The performance decreases slightly
as the ratio of the batch size to the number of classes increases. To
demonstrate the efficacy of the block-random algorithm for
exascale-type simulations, we used a \gls{dns} simulation of turbulent
channel flow to construct a \gls{les} \gls{sgs} model. The model
constructed using the block-random algorithm performed significantly
better than the block-unshuffled learning model and is within $8\%$ of
the fully shuffled model. Using the Jensen-Shannon divergence metric,
we analyzed the characteristics of the batches used by the \gls{sgd}
to inform a criteria for successfully constructing a \gls{dnn} model
for distributed heterogeneous data.

This work --- including neural network models, analysis scripts,
Jupyter notebooks, and figures --- can be publicly accessed at the
project's GitHub
page.\footnote{\url{https://github.com/NREL/block-random}} Traditional
machine learning algorithms were implemented through
scikit-learn~\cite{Pedregosa2011} and the deep learning algorithms
through Keras~\cite{Chollet2015} with the TensorFlow
backend~\cite{tensorflow2015-whitepaper}.

\section*{Acknowledgments}
This work was authored in part by the National Renewable Energy Laboratory, operated by Alliance for Sustainable Energy, LLC, for the U.S. Department of Energy (DOE) under Contract No. DE-AC36-08GO28308. Funding provided by U.S. Department of Energy Office of Science and National Nuclear Security Administration. The views expressed in the article do not necessarily represent the views of the DOE or the U.S. Government. The U.S. Government retains and the publisher, by accepting the article for publication, acknowledges that the U.S. Government retains a nonexclusive, paid-up, irrevocable, worldwide license to publish or reproduce the published form of this work, or allow others to do so, for U.S. Government purposes.

This research was supported by the Exascale Computing Project (ECP), Project Number: 17-SC-20-SC, a collaborative effort of two DOE organizations -- the Office of Science and the National Nuclear Security Administration -- responsible for the planning and preparation of a capable exascale ecosystem -- including software, applications, hardware, advanced system engineering, and early testbed platforms -- to support the nation's exascale computing imperative.

\section*{References}

\bibliography{library}

\end{document}